# Generalisation of language and knowledge models for corpus analysis

Anton Loss (avl@gmx.co.uk)

ABSTRACT. This paper takes new look on language and knowledge modelling for corpus linguistics. Using ideas of Chaitin, a line of argument is made against language/knowledge separation in Natural Language Processing. A simplistic model, that generalises approaches to language and knowledge, is proposed. One of hypothetical consequences of this model is Strong AI.

## Introduction

As Plungyan(2009) noticed, corpus linguistics made it possible to treat language impartially. Corpus, which is a large collection of human written texts, gives opportunity to depart from subjective imposing of what is *right* and *wrong* in both knowledge and language. When something is illogical, or syntactically incorrect, but it is found in corpus – then it is not discarded. This paper shows that when treating language in such impartial manner, then there is no major difference between syntax and knowledge models.

## Reasons for syntax separation today

Complete separation of syntax from analysis of meaning or knowledge was done in the middle of 20 century (Plungyan, 2009). Possibly strongest reason for separation was subconscious feeling that many grammatical forms can be associated with the same meaning. It is impossible though to definitely prove that, for instance, "This is Mary's pony" and "This pony belongs to Mary" mean *exactly* the same.

Today language syntax is modelled separate from knowledge and logic. Two popular approaches will be used here as examples of each type of model. Analysis of language (syntax) will be represented by Chomsky's generative grammar. Analysis of knowledge will be represented by a programming language Prolog. It will be shown that both methods are individual cases of more general approach. A brief description of how Prolog and Generative Grammar are viewed in this paper follows.

### Prolog

Prolog is computer implementation of formal logic. It can be reduced to the following:

Several true statements, such as "Mary likes ponies" or "Mary is a girl", are given. These statements are put into Prolog syntax:

> LIKES (MARY , PONY)

> GIRL(MARY)

Logical statements can also be given. It is possible to define logic such as "any girl likes ponies".

> LIKES(X, PONY) :- GIRL(X)

After finite amount of statements is given to Prolog it is possible to obtain a list of all valid statements, which could be derived from them. Following statements are given:

> **GIRL(LINDA)**
>
> **GIRL(MARY)**
>
> **LIKES(X, PONY) :- GIRL(X)**

Prolog will produce following:

> **GIRL(LINDA)**
>
> **GIRL(MARY)**
>
> **LIKES (MARY , PONY)**
>
> **LIKES(LINDA, PONY)**

## Generative Grammar
(For simplicity only context free grammar is considered here.)

Generative grammar can be essentially reduced to generating stings by unconditional substitution. The starting symbol(**S**) and rules of substitution are defined, and from them strings are generated. For instance rules can have following form:

> **S → SS,**
> **S → A,**
> **S → B,**

This will produce strings as:

> **S**
>
> **SS**
>
> **SSS**
>
> **SSA**
>
> **SBA**
>
> ...

In this particular case strings of all possible combinations of "S", "a" and "b" are produced.

## Chaitin argument against language/knowledge separation
Kolmogorov-Chaitin complexity of a string is defined relative to the length of a shortest program that can produce such a string (Chaitin 1987).

In his later works Chaitin(2006) argues that comprehension is essentially the process of compression. So for instance to understand the string "abababababababababababab" would be to describe it as "12 times ab". As second string is only 11 symbols long, while original is 26, Chaitin (2006) argues that to produce the second string is same as to understand the first one.

If instead of one string, a set of stings is described, then this idea can relate to understanding of corpus.

### *Complexity in Prolog*

(In following paragraphs term 'corpus' is slightly misused. And it means a set of statements of any sort. 'National corpus' is used instead to refer to a large collection of human created texts.)

It is possible treat corpus as collection of separate sentences. Those sentences are compressed to produce a program that produces those sentences back. Let's call this program Compressed Corpus (CC). Length of CC should be shorter or equal to length of original corpus. Addition or removal of sentences from original corpus will be changing the length of CC.

It is possible to show that, at least for the case of Prolog, addition of some sentences will increase length of CC, while addition of others will decrease it. Consider that corpus consists purely of Prolog statements, and that CC is also written in Prolog. For instance:

 FATHER_CHILD(TOM, SALLY).
 FATHER_CHILD(TOM, ERICA).
 FATHER_CHILD(TOM, JAMES).
 FATHER_CHILD(TOM, POLY).

 SIBLING(SALLY, ERICA).
 SIBLING(SALLY, JAMES).
 ...
 SIBLING(JAMES, SALLY).
 SIBLING(ERICA, SALLY).

If in these statements all possible siblings are identified, then this can be compressed into the following CC:

 FATHER_CHILD(TOM, SALLY).
 FATHER_CHILD(TOM, ERICA).
 FATHER_CHILD(TOM, JAMES).
 FATHER_CHILD(TOM, POLY).

 SIBLING(X, Y) :- FATHER_CHILD(TOM, X), FATHER_CHILD(TOM, Y).

The second program will produce exactly same statements as the first one. The list of siblings was simply replaced with a rule, to generate such list.

If one sibling relation is removed from original corpus, it is no longer possible to use old CC to produce it, as old CC will produce one extra relation which is no longer part of the corpus. So after removal of one relation, new CC will become longer. Adding that statement back will, in turn, decrease size of CC.

If a statement without any connections to other statements is added, then CC would increase in length. It will decrease when it is removed again.

This shows compression aspect to the formal logic and Prolog in particular - many valid statements are compressed into fewer statements (some of which are rules).

## *CC limited in size*

Assume that size of CC is limited. In this case there would be times when it won't be possible to compress all statements of original corpus. So trade-offs would have to be made to fit CC into certain size: either not all original statements will be compressed, or some extra statements will be compressed. Let's call those two tradeoffs *Completeness* and *Accuracy*.

A CC limited in size balances between *Accuracy* and *Completeness*. Two extremes could be identified. Imagine size of the CC is very limited. Then there would be two ways to compress a large corpus:

**Near 100% accuracy with near 0% completeness**

> This would be several statements from original corpus in their original form. Except for those few all other statements are not compressed and lost.

**Near 0% accuracy with near 100% completeness**

> This would be when near all possible statements are generated without regards to original corpus. So all original statements will be preserved, but it will be impossible to distinguish them from other random sequences of words. It is not obvious that this extreme compression is as useless as the other one (Borges, 1941).

## *Complexity in Generative Grammar*

Generative grammar uses what appears to be different, but essentially the same approach. As with Prolog it is possible to encode sentences "as is" with it:

> S → TOM IS SALLY'S FATHER
> S → TOM IS ERICA'S FATHER
> S → TOM IS JAMES'S FATHER
>
> S → SALLY AND ERICA ARE SIBLINGS
> S → SALLY AND JAMES ARE SIBLINGS
> …

This would reproduce initial statements without any compression. On the other hand it is possible to achieve the same with different rules.

> S → TOM IS C FATHER
> C → SALLY
> C → ERICA
> C → JAMES
>
> S → C AND C ARE SIBLINGS

If a National Corpus is compressed correctly with Generative Grammar, then theoretically resulted CC should have near 100% completeness, with measurable positive accuracy. The fact that Generative Grammar usually produces infinite number of statements can be overcome by limiting sentence length to the longest sentence in National Corpus.

## *Prolog and Generative Grammar compared*

In essence both Prolog, and Generative Grammar use substitution for compression. In a way they both work towards understanding of a language. Their success of understanding language can be compared to each other. Assume that there is some simple function that translates Prolog statements into human readable, grammatically correct form.

The following graph demonstrates how Prolog and Generative Grammar can be compared on the chart. Y-axis measures what percentage of sentences in national corpus can be produces by particular method. X-axis measures percentage of sentences produced by the method that can be found in national corpus. So Y-axis is *completeness* and X-axis is *accuracy*.

Set a maximum program size, and try to find such Prolog program and Generative Grammar rules no bigger than that size, that will be generating corpus sentences as *completely* and *accurately* as possible. Then reduce allowed size and try to repeat this process. It is possible to plot this on graph.

- Line *p* indicates Prolog.
- Line *g* indicates Generative Grammar.
  The direction of the lines indicate decreasing limit of the program.

  Extra points are added for comparison:
- *a* is half of the sentences from national corpus
- *b* is half of national corpus sentences with addition of same amount of sentences not found in national corpus
- *c* is national corpus itself

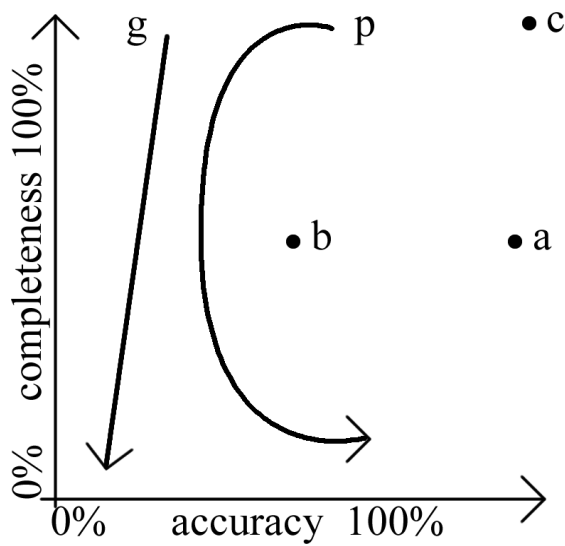

Set of statements produced by particular method is called **M**. Set of statements in national corpus is **C**.

$$\text{ACCURACY} = |(M \cap C)|/|M|$$

$$\text{COMPLETENESS} = |(M \cap C)|/|C|$$

Last graph demonstrates how different methods can be compared. Actual lines might look differently. In fact for each size limit there won't be just one point, but a number of points, most efficient of which can be connected into a line.

Let's say that limit is fixed for each compression method at some value. For each fixed *completeness* and size there would program with maximum *accuracy*, and vice-versa. So it is possible to plot *completeness* against *accuracy* for fixed size. Once again, following graph presents only concept of comparing Prolog and Generative Grammar at fixed program size, actual lines could look much different.

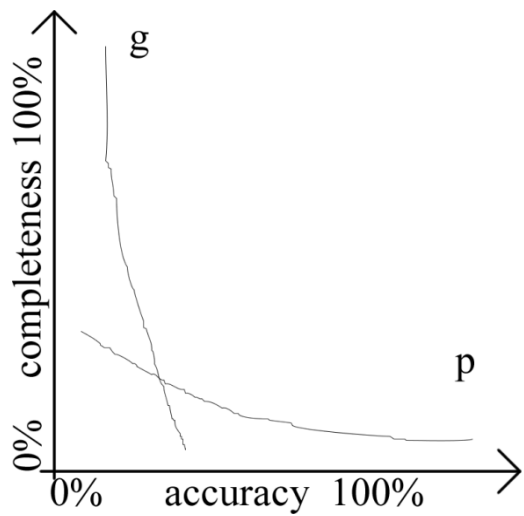

Both Prolog and Generative Grammar can be called methods of *loose compression* (LC). LC can be related to understanding of any structure that can be expressed through separate statements that consist of finite alphabet. LC is similar to finding some order in a set of such statements in order to compress it into a smaller set of statements, which in turn can be used to *loosely* produce back original set.

It seems that many methods for LC could exist. LC should be able to vary between *accuracy*, *completeness* and *size*. This means that good LC method should be capable of storing statements "as is", to produce 100% *accuracy* and 100% *completeness* when size of CC is limited to the size of original set. It should be flexible to maximise its *accuracy* when *completeness* and size are set, and vice-versa.

An attempt to find most general and flexible LC method is presented below.

## Bracket Compression

Bracket Compression (BC) is a general *loose compression* method. It facilitates compression of strings of any kind in any form: sentence parts, sentences, paragraphs, and texts. BC syntax consists only of well formed brackets and natural language words (or symbols of any kind) between them.

Only rule for processing statements with brackets is to replace brackets with the ending of another statement that starts with the content of the brackets and does not contain brackets itself. For example:

>   [GIRL] LIKES PONIES

>   GIRL MARY

This will produce:

>   [GIRL] LIKES PONIES
>
>   GIRL MARY
>
>   MARY LIKES PONIES

If the brackets contain a full existing statement then they are removed (replaced with nothing). If there are several brackets containing same fragment, they all are replaced the same way. This simulates effect of variables without the need for actual variables. If brackets are empty they are replaced with any complete statement, all empty brackets are replaced the same way.

>   NAME MARY
>
>   MARY IS A GIRL
>
>   NAME TOM
>
>   TOM IS A BOY
>
>   [NAME] LIKES PONIES [[NAME] IS A GIRL]

Among other statements this will produce

>   MARY LIKES PONIES [MARY IS A GIRL]
>
>   TOM LIKES PONIES [TOM IS A GIRL]

Because there is no statement to replace "[Tom is a girl]", only one sentence without brackets will be generated:

>   MARY LIKES PONIES

It is possible to simulate Chomsky context-free grammar with BC. Take for instance grammar with following production rules

>   S → ASA,
>   S → BSB,
>   S → E,

The corresponding BC would be:

>   S→A[S→]A
>
>   S→B[S→]B
>
>   S→

This time "→" was kept, but it has no special meaning in BC, it could've been any other symbol; except for brackets, meaning of everything else is defined within the compression itself.

Because multiple brackets with same content in BC are processed as variables, multiple non-terminals are numbered. Consider following Generative Grammar rule:

> S→SS

This can be achieved with two BC statements.

> S→[S→][S1→]
>
> S1→[S→]

It is also possible to imitate Prolog statements:

> GIRL (MARY)
>
> LIKES(X, PONIES) :- GIRL(X)

This becomes:

> GIRL MARY
>
> [GIRL] LIKES PONIES

Alternatively this can become:

> GIRL MARY
>
> MARY
>
> [] LIKES PONIES [GIRL [] ]

Most statements can be translated from Prolog to BC.

### *Advantages of BC for Corpus analysis*
- It is possible to establish connection between form and meaning. For instance if text is about monarch, plural pronoun might be used instead of singular.
- Instead of trying to model the meaning of the sentence, its connection to other sentences is captured, which makes work of scientist more impartial.
- It is possible to express illusive logic such as "Mary likes some ponies" also means "She is probably not fond of some other ponies".
- Meaning of the word or word combinations within some specific context could be captured. Some phrases could for instance mean slightly different thing in some particular context.
- It is possible to find a place of negative examples in the language. Starred sentences (examples of incorrect sentences) are now part of the language, as they are part of national corpuses. Other models would fail to address them by definition.
- It is possible to analyze information that is not considered a part of language, but plays some ambiguous role, for instance emoticons.
- BC has only one type of statements and minimum artificial syntax and logic to it. This means that the logic expressed with it, is the logic that comes from the language itself.

# Formal definition of Bracket Compression

Bracket Compression (BC) is a simplified compression method. All statements are of the same form, and can be generated with the following context-free grammar:

>G = ({S},{WORDS, [, ] },S,P)
>
>S → S S
>
>S → [S]
>
>S → WORD1 | WORD2 | WORD3 | ...

Every set of N statements in BC can be expanded into N or more statements. The expansion is made by replacement of the words in the brackets, with the ending of the statement that starts with these words, and does not contain brackets itself. Statements with brackets can be compared to logical statements in Prolog. In Generative Grammar they can be compared to rules, or strings with non-terminals.

>*A [B]*
>
>*B C*

Can be expanded to

>*A [B]*
>
>*B C*
>
>*<u>A C</u>*

If several brackets contain the same sequence of words they are replaced the same way:

>*A [B] [B]*
>
>*B C*
>
>*B D*

This can be expanded to

>*A [B] [B]*
>
>*B C*
>
>*B D*
>
>*<u>A C C</u>*
>
>*<u>A D D</u>*

Notice that neither **A C D** nor **A D C** can be obtained from the original set.

Empty brackets are replaced with a complete existing statement. All empty brackets within the statement are replaced the same way.

*A [] [ [] C]*

*B*

Produces:

*A[] [ [] C]*

*B*

*A B [B C]*

## Usage examples

One possible application would be to find a rule that connects two sentences. And then generate large amount of such pairs and look if there are texts where they stand together.

Sufficient amount of trials might help discover connections between seemingly unconnected sentences. Or show connection between sentences that seem to be connected. Such rule can have following form:

*[NAME] WAS WEARING [ACCESSORY] [SOME OTHER SENTENCES]*

*[PR] TOOK OFF [ACCESSORY] [PRONOUN OF [NAME] IS [PR]]*

This statement parts mean the following:

*[[NAME] WAS WEARING [ACCESSORY]]* **–** produces first sentence. Name generates name. Accessory is an accessory.

*Mary was wearing necklace*

*[SOME OTHER SENTENCES]* **–** produces sentences (or words) in between. In the implementation no actual sentences are generated. This just denotes that something can be between first and second sentence.

*[PR] TOOK OFF [ACCESSORY]* **–** Produces second sentence, where Pr is pronoun, and accessory is the same accessory.

*She took off necklace*

*[PRONOUN OF [NAME] IS [PR]]* **–** matches a name with the pronoun used in the main sentence.

*Pronoun of Mary is she*

This would find a text where "Mary was wearing necklace" is followed by "She took off necklace" (articles are missing for simplification). Also text with "He took off tie" after "John was wearing a tie" would be found.

This demonstrates everyday kind of logic – "what is taken off was once worn". This kind of logic is otherwise is hard to capture. Also this captures that when something is worn, the person who is wearing it is addressed by name, and when the person takes it off, then the person is addressed by a pronoun. This is probably not true – but some relations of this sort might exist.

A simple search in Google shows that this method indeed can be applied to real texts, for instance search for:

>  "she was wearing" "she took off"

This finds several texts, where item worn is the same item that is taken off.

### Addition statements
With BC it is possible to recursively specify addition.

AFTER 0 IS 1
AFTER 1 IS 2
AFTER 2 IS 3

etc..

NUMBER 0
NUMBER [AFTER [NUMBER] IS]
BEFORE [NUMBER] IS [ANOTHER NUMBER] [AFTER [ANOTHER NUMBER] IS [NUMBER]]
ANOTHER NUMBER [NUMBER]
[NUMBER] + 0 = [NUMBER]
[NUMBER] + [ANOTHER NUMBER] = [[AFTER [NUMBER] IS] + [BEFORE [ANOTHER NUMBER] IS]]

This would be sufficient to recursively generate all "n + m = k" statements. It is very unlikely that in natural language such amount of recursion would be found, but this could be suitable for generating scientific part of the corpus.

### Make compression with NOT

> TOM IS A GIRL, NOT!
>
> MARY IS A GIRL
>
> NAME MARY
>
> NAME TOM
>
> [NAME] LIKES PONIES [[NAME] IS A GIRL]

This would produce a negative statement without logic specified.

> MARY LIKES PONIES
>
> TOM LIKES PONIES, NOT!

This might help to better understand double negatives in some languages. This also shows the possibility that logic needs not to be artificially modelled. Also there is some evidence in the other NLP (neuro-linguistic programming) that humans are easily confused with multiple negations (Erickson, 1976). This could be related the possibility that human do not have "logical negation" in Computer Science sense.

### Text compression
With BC it is theoretically possible to compress a whole text. For instance:

> ONCE UPON A TIME THERE LIVED [NAME OF THE [HERO]]. [PRONOUN OF [HERO]] WAS [POSITIVE QUALITY OF [GENDER OF [HERO]]]. ONE DAY [NAME OF THE [HERO]] [OUTSTANDING ACTION]. *etc..*

It is possible to represent in this form any particular fairytale. Theoretically it is possible to find a template for all fairytales, though that is probably almost as hard as to find a common template for all texts.

### AI
It might be possible to define AI from BC. AI would be such a compression of national corpus that its size is reasonable small, its accuracy is reasonably low, and its completeness is reasonably high. It might be possible to estimate those parameters without actual construction of AI. For instance they could look like this:

> Size: 100MB
>
> Accuracy: 0.00002%
>
> Completeness: 99.98%

This would give at least some idea to scientists what AI machine might look like. If this numbers could've been found, it would be possible to argue that this is an improvement over Turing Test.

Accuracy is set so low because it can be thought that there is some bigger, theoretical corpus, which includes all conceivable texts (including dialogs during Turing Test). Real corpus is only smallest fraction of all possibilities, though it should be sufficient to give enough evidence for creating correct rules. Accuracy is probably many magnitudes smaller in reality that estimated above.

### Text generating
Compressing example sentences with low *accuracy* could be used to produce pseudo-random texts, as an alternative to Markov Chains.

### Partial understanding
This model can provide partial understanding of the language, some logical observations "out of the apparently monolithic problem of natural language understanding" (Abney, 1996). It will be possible to statistically test those relations using national corpuses.

## Possible developments

### Functions
It might be sensible to add system functions to BC. Comparison function would aid practical compression and help increase accuracy. This function would be automatically generated by the system hosting BC and would iterate through all possible words and sentences telling which are different. Produced rules could have following form:

> DIFFERENT A FROM B
>
> DIFFERENT A FROM C

DIFFERENT B FROM C D

etc.

Though it is possible in some cases to compress even this, it seems to be quite complex.

## Conclusion

In this paper it was analysed that in fact two common approaches of analysing language and knowledge are variations of a more common approach. This common approach was defined. This approach uses language itself as the basis, and establishes connections between real language sentences, instead of modelling it. This approach might gain popularity in modern linguistics due to appearance of corpuses, as they make it possible to utilise this method. Also this model provides alternative view on Strong AI.